# ICP Algorithm: Theory, Practice And Its SLAM-oriented Taxonomy

Hao Bai [0000-0001-9723-7490]

University of Illinois Urbana-Champaign, UIUC
Champaign, Illinois, USA
`Haob2@illinois.edu`

**Abstract.** The Iterative Closest Point (ICP) algorithm is one of the most important algorithms for geometric alignment of three-dimensional surface registration, which is frequently used in computer vision tasks, including the Simultaneous Localization And Mapping (SLAM) tasks. In this paper, we illustrate the theoretical principles of the ICP algorithm, how it can be used in surface registration tasks, and the traditional taxonomy of the variants of the ICP algorithm. As SLAM is becoming a popular topic, we also introduce a SLAM-oriented taxonomy of the ICP algorithm, based on the characteristics of each type of SLAM task, including whether the SLAM task is online or not and whether the landmarks are present as features in the SLAM task. We make a synthesis of each type of SLAM task by comparing several up-to-date research papers and analyzing their implementation details.

## 1 Introduction

The Iterative Closest Point (ICP) algorithm is becoming more and more popular for heuristically registering three-dimensional surfaces based on geometry without any help from depth or structural information. Typically, the ICP algorithm is used for aligning two surfaces (usually represented by mesh grids), which makes it popular as a built-in modeling functionality of 3D scanners and perceptual cameras. The ICP algorithm was first invented by Besl *et al.* [1] and has continuously been developed by researchers until now.

Earlier research has explored different types of ICP algorithms, like changing the point-to-point registration metric to a point-to-line [2] or point-to-surface [3] one, developing a coarse-to-fine multi-resolution approach to accelerate the ICP algorithm in a high-resolution context [4], and utilizing more extra data like hue data when applying the ICP algorithm [5].

Earlier work has also been done to compare and synthesize variants of ICP algorithms. Szymon *et al.* discussed efficient variants of the ICP algorithm [6], Pomerleau *et al.* compared the performance of variants of the ICP algorithm using real-world datasets [7], and Erza *et al.* analyzed the rigorous upper and lower bound of variants of ICP algorithms theoretically [8].

However, despite the quick growth of Simultaneous Localization And Mapping (SLAM) techniques [9, 10], the estimation and application of the ICP algorithm inside this certain representative field are not covered by most of the literature. Thus, we propose a brand-new SLAM-oriented taxonomy of ICP algorithms and their applications for a handier use of selecting a suitable ICP algorithm for SLAM researchers.

## 2 Related Works

### 2.1 SLAM

By official definition, SLAM is the process for a mobile robot to build a map of the environment and use this map to compute its location simultaneously [10]. For a more detailed illustration, SLAM is a method for an autonomous vehicle to start in an unknown location in an unknown environment and then incrementally build a map to compute the absolute vehicle location [11]. A graphical explanation of SLAM based on a point cloud approach is also shown (Figure 1). With the fast development of 3D computer vision, SLAM is playing a more and more important role in both academia and industry. Typical applications of SLAM include analyzing proposed designs of an aerospace manufacturing facility, evaluating the throughput of a new pharmaceutical production line [12], self-driving automobiles [13], augmented reality, and various kinds of other fields concerning robotic movements.

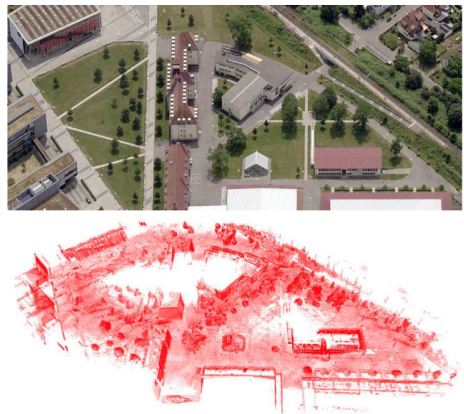

**Fig. 1.** Top: satellite image acquired at the University of Freiburg. **Bottom**: the relative point cloud map (with courtesy of Kai M. Wurm). Note that both pictures are from [14].

Up to now, there have been lots of implementation variants of SLAM, including a versatile and accurate monocular SLAM system called ORB-SLAM [15], a real-time single-camera SLAM called Mono-SLAM [16], a large-scale direct monocular SLAM called LSD SLAM [17], and so on. However, all these variants of SLAM follow the same basic methodology of a SLAM system, as listed below[1] (graphically illustrated by Figure 2).

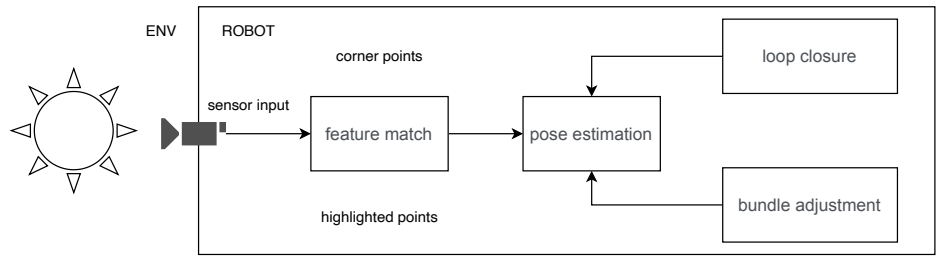

**Fig. 2.** A graphical explanation of the basic SLAM system.

- **Sensor input.** The sensor acts as the only input of the SLAM system. The sensor on the robot (usually RGBD camera or LIDAR) sends real-time information to the data processor.
- **Feature match.** The feature match step processes the data input and figures out whether the features in the input data match certain kinds of features pre-defined in the robot memory, like corner points and highlighted points. Some research focuses on amelioration of this step like variants of Gaussian algorithm applied [18] and the invention of the ORB algorithm [15].
- **Pose estimation.** This part estimates the movement of the input data between continuous frames with the help of matched features and utilizes methods like eliminating outliers and denoising to enhance the performance of pose estimation.
- **Loop closure.** The loop closure part determines whether the robot has been to a certain position before, which is equivalent to examining whether the machine has achieved a closure, to help with a more accurate pose estimation.
- **Bundle adjustment.** The bundle adjustment part minimizes reprojection error of the sensor, which requires a solution to least-squares problems, to help with a more accurate pose estimation.

Inside all parts above, there are multiple places where we can utilize a 3D surface registration for boosting the performance of the SLAM system. The most used improvements for 3D shape registration are the pose estimation and loop closure part. For example, in the pose estimation part, 3D surface registration can be used to align the partial overlaps between the point clouds in two continuous frames [19], and in the loop closure part, 3D surface registration can be used for registering the pair of the images where the closure is achieved [20]. Thus, we take another subsection for illustrating both the ideology and methodology of basic 3D surface registration.

---

[1] In the industry, there is also a jargon called VIO (VIsual Odometry). Intuitive speaking, the only difference between SLAM and VIO is that, VIO does not have the pose estimation and build adjustment part while SLAM has them.

## 2.2 3D Surface Registration

3D surface registration is the process that aligns three-dimensional data from different viewpoints or at continuous frames, and general applications of 3D surface registration include reconciliation of surfaces for obtaining a more complete scene [21, 22] and estimation of motion errors for a moving camera. The graphical illustration of a 3D registration process is shown (Figure 3).

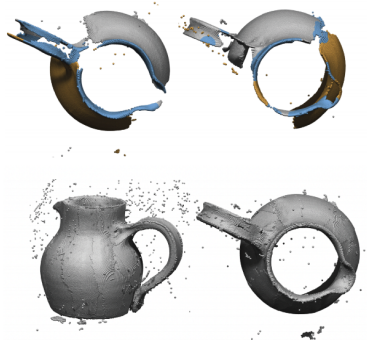

**Fig. 3.** Process of 3D surface registration. **Top**: input data. **Bottom**: the registered and reconstructed result. Note that the graph is made by [23].

With the rapid development of concerning data structures like polygon mesh and point cloud, 3D surface registration is becoming more and more popular in fields like medical imaging [24], object recognition [25], data fusion [26], and most commonly, SLAM [27-30].

However, different applications of surface registration led to variants of surface registration. There are a variety of types of surface registration, and different types of surface registration require different algorithms to fulfill the task, as listed below.

For tasks with known association data, the association of the input pair of images is given in the input data [31]. In the context of surface registration, the translational and rotational shift between the two images is included in the input data. For this kind of task, no heuristic algorithm is needed, and it's already proved to be solvable by an SVD-based alignment[2] in a closed-form.

The methodology of surface registration with known correspondences is mathematically illustrated by Formula 1, where source point cloud $S = \{s_1, s_2, ..., s_I\}$, destination point cloud $D = \{d_1, d_2, ..., d_J\}$, and the correspondence $\mathcal{C}(s_i) = \{(i,j)\}$ are given, and the translation vector $t$ and rotation matrix $R$ are to be determined, to minimize the distance sum between all the points on the point clouds. The operation is then marked with the operator $\mathcal{F}$.

$$\mathcal{F}(\mathcal{C}(S), S, D) = \min_{R,t} \sum_{i,j \in C} \|d_i - R \cdot s_j - t\|_2^2 \quad (1)$$

---

[2] SVD is the abbreviation for "Singular Value Decomposition" in numerical optimization.

For tasks with unknown association data, we need to "guess" the correspondences (association between the two surfaces) with a certain strategy for repeated times, and each time after guessing the correspondences the problem becomes tasks with known association data again. Thus, no direct solution is provided, and the only way to solve is a heuristic approach. There are multiple algorithms for this kind of task, like numeric Newton method [32], dynamic genetic algorithm [33], and most commonly, variants of the ICP algorithm [3, 34, 35] for SLAM system for especially LIDAR or RGBD data, which we're mainly concerned with for our research.

In the next part, we're going to dive deep into the basics of the ICP algorithm with both its ideologies and methodologies and the previous taxonomy of the ICP algorithm by other researchers.

## 3 Basic ICP Algorithm and Previous Taxonomy

### 3.1 Basic ICP Algorithm

The basic ideology of the ICP algorithm is simple: if the correspondence between the point clouds is known, we can always get the transformation from the source point cloud to the reference point cloud, as it's the same as the case of solving tasks with known association data. The problem is that the correspondence we guess is not exact, so we need to iterate the guess-transform process several times using a `while` loop.

The methodology of the basic ICP algorithm is divided into 3 parts. First, the ICP algorithm is fed with input data, and it sets up an initial guess of the possible association of the pair of surfaces. Second, the ICP algorithm iterates many times, within each time it makes some movements to the source surface so that it becomes closer to the destination surface. At last, when approached the threshold of convergence, the ICP algorithm stops iteration and returns the output data including the translational vector and the rotational matrix.

In more detail, the algorithm is illustrated by Algorithm 1 below, where source point cloud $S = \{s_1, s_2, ..., s_I\}$, destination point cloud $D = \{d_1, d_2, ..., d_J\}$ and convergence threshold $\theta_0$ are given, the center of mass operator $\mathcal{M}$, the known-correspondence surface registration solution operator $\mathcal{F}$ and the error computed for a certain epoch of iteration $\epsilon$, the correspondence $\mathcal{C} = \{(i,j)\}$, the normal transformation operator $\mathcal{T}$ and transformation with arguments operator $\mathcal{T}_{R,t}$ can be utilized.

The graphical illustration of the ICP algorithm is also shown in Figure 3, and the `C++` implementation of the basic ICP algorithm is available on the GitHub Page[3].

---

[3] You can also access manually by URL https://github.com/BiEchi/IterativeClosestPoint. Note that this repository is under GPL license. You can start by reading the instructions in `README.md`.

**Algorithm 1:** Basic ICP Algorithm

**Data:** Source point cloud $S$ and destination point cloud $D$, convergence threshold $\theta_0$.
**Result:** The transformed source point cloud $S'$.

1 $S' := T(S) \ s.t. \ \mathcal{M}(S) = \mathcal{M}(D)$;
2 $\epsilon := \infty$;
3 **while** *($\epsilon > \theta_0$ **and** $\epsilon$ has decreased) **or** $\epsilon = \infty$* **do**
4     **for** $s_i \in S$ **do**
5         $\mathcal{C}(\mathbf{s}_i) := (\mathbf{d}_j, \mathbf{s}_i) \ s.t. \ \|(\mathbf{d}_j, \mathbf{s}_i)\|^2 \to min$;
6     **end**
7     $S = S'$;
8     $\mathbf{R}, \mathbf{t} := \mathcal{F}(\mathcal{C}(S), S, D)$;
9     $S' := T_{\mathbf{R},\mathbf{t}}(S)$;
10     $\epsilon := \frac{1}{|I|} \sum_{i \in I} (\mathbf{R}\mathbf{s}_i + \mathbf{t} - \mathbf{d}_j)^2$;
11 **end**
12 **return** $X, Y'$;

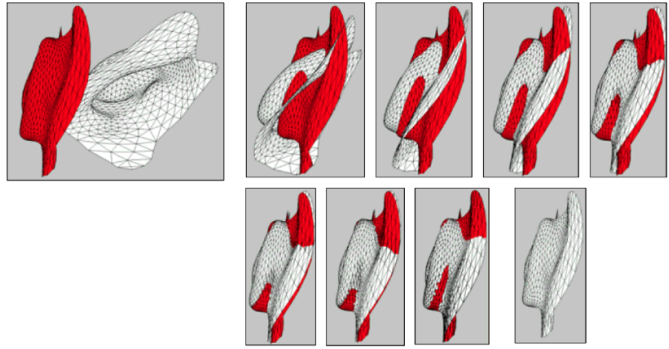

**Fig. 4.** The data output of the basic ICP algorithm with the iteration goes on. The first step is transforming the source mesh so that the centers of mass of the two meshes come together, by Technion at https://www.cs.technion.ac.il/~cs236329/tutorials/ICP.pdf.

### 3.2 Previous ICP Taxonomy

As mentioned in the Introduction part of this paper, there have already been kinds of literature concerning taxonomies of the ICP algorithm, because of the wide application and variants of the ICP algorithm. One of the most famous taxonomies of the ICP algorithm is the essay *Efficient Variants of the ICP Algorithm* by Szymon Rusinkiewicz and Marc Levoy [6]. In their research, the classification of the variants can be limited to six stages of the algorithm, i.e., the point selection stage, the correspondence matching stage, the correspondence confidence decision stage, rejecting outlier stage, the error metric assignment stage, and the error metric minimization stage. According to these taxonomy criteria, variants of the ICP algorithm are classified into different types. For each stage, there are three to five high-performance variants compared, following a structure-oriented taxonomy methodology. Our re-search follows an application-oriented method that, we firstly determine a taxonomy of SLAM, then use this taxonomy for deriving the taxonomy of the ICP algorithm, which mainly focuses on the application of ICP algorithm in SLAM.

## 4  Selection of ICP and Bayes Filter

Before classifying ICP variants in the SLAM system, we need to determine whether we should use the ICP algorithm at all. There are two main algorithms in SLAM to perform the data processing step, i.e., using state estimation approach with Bayesian theorem or using scan matching approach with ICP algorithm [36]. Bayesian theorems are well established, but if the environment is large, the number of landmarks used for calculation increases and calculation becomes difficult. ICP algorithm, instead, estimates the movement amount of the robot and avoids this problem.

For SLAM, the state estimation task [37] is equivalent to estimating where the robot is in the environment, given all information about where it was, and what command it was given [38, 39]. The mathematical illustration of state estimation is shown by Formula 2, where $\theta$ represents the state estimation task, $x_t$ stands for the state or position of the robot at time $t$, $z_{1:t}$ means all the state information before time $t$, and $u_{1:t}$ represents the sequence of commands we give to the robot before time $t$.

$$\theta = p(x_t | z_{1:t}, u_{1:t}) \qquad (2)$$

The recursive Bayes filter is a recursive framework for state estimation tasks, which gains its name "recursive" because we use not only the new distribution we get but also reuse the previous distribution we have, to gain a more precise understanding of where the robot is [40]. The Bayes filter solution can be expressed by Formula 3, where $\mu(x_t)$ means the degree of belief at the position $x_t$ at time $t$, $\eta$ stands for the normalization factor, $p(x_t | u_t, x_{t-1})$ means the probability of the robot appearing at position $x_t$ given the command at $u_t$ and its position at time $t - 1$.

$$\mu(x_t) = \eta \cdot p(z_t | x_t) \cdot \int p(x_t | u_t, x_{t-1}) \cdot \mu(x_{t-1}) \, dx_{t-1} \qquad (3)$$

The recursive Bayes filter has many variants of implementations used in SLAM, like Kalman filter, extended Kalman filter (EKF) [41], and particle filter [42]. Kalman filter is the Bayes filter for the Gaussian linear case, and it has 2 steps: prediction and correction. EKF is the extension of Kalman filter for non-linear tasks, and the particle filter is mostly the same as the Karmen filter but relaxes some limitations that Kalman filter does.

Thus, it's also important to select from the ICP algorithm and the Bayes filter in the data processing part of the SLAM system. Only after the decision of using the ICP algorithm should one try to find out a suitable ICP variant for his SLAM system. The next part concerns what kind of ICP algorithm to use if we decide to use it.

# 5 SLAM-oriented ICP Taxonomy

As already mentioned in the Related Works part of this paper, there is a large amount of research concerning SLAM variants, e.g., ORB-SLAM, Mono-SLAM, and LSD-SLAM. Different types of SLAM systems have different requirements for accuracy, power consumption, time efficiency, and data processing patterns. We classify the types of SLAMs in all the criteria shown below.

## 5.1 Online and Offline SLAM

There are two types of SLAM, online and offline. In the field of machine learning, online learning means that the learning process is going on when the data is coming in, and offline learning means the dataset is already static for learning. Following the same rule, online SLAM (or filtering SLAM) means that when the robot is moving, it can not only detect its vicinity but also use this input for analysis of localization and mapping [43, 44], while offline SLAM (or smoothing SLAM) means that when the robot is moving, it stores the data detected as input locally, and after collecting all data, it begins to analyze the localization and mapping stuff [45, 46].

In the SLAM industry, online methods and offline methods have different applications. According to Thrun *et al.*, offline methods provide a possibility to revisit all data, instead of discarding them [47], which improves the data maneuverability of offline methods. On the other side, online SLAM is time-sensitive, because the data is streamed and the system needs quick a response. Recently, research is focusing on achieving as good results using online methods as those using offline methods. We analyze the characteristics of online and offline SLAM individually to perform an ICP taxonomy below.

For an online SLAM system, time is the most important ingredient for data processing, so a high-efficiency ICP algorithm is needed [6, 48]. In this kind of SLAM, the time consumption should be as low as possible. As the data input is one-directional, the algorithm can't get global data, which means there is an inevitable loss in precision. Some outstanding work on online SLAM methods since 2000 is included below. Sharf *et al.* (2006) introduced an interactive technique for easy mesh composition with a newly proposed ICP algorithm called Soft-ICP algorithm, which enhances registration to elastic transformations that account for simultaneous global positioning and local blending of the objects [49], which can be used for SLAM as well. Holz *et al.* (2010) proposed an ICP algorithm without preprocessing step, that allows to determine which points in a newly acquired range image are already contained in the point model. This algorithm only adds points providing new information [50].

For an offline SLAM system, precision is the most important factor. In the offline model, the main task of the physical robot is to store and pre-process data for later analysis, so the time to offline methods is not as crucial as online methods. In this field, there are also lots of new techniques since 2000. Moosmann *et al.* (2011) proposed a Velodyne SLAM model, where a filtering step is applied to refine the produced map offline, and ICP only twice - before and afterward [51]. This research proposes that de-skewing should be employed at each iteration of the ICP algorithm. According to the results of this research, de-skewing only twice is sufficient for the offline model. De-skewing is especially useful for offline mthods because the algorithm has access to global data. In another research, Dai *et al.* (2019) developed an offline coarse-to-fine SLAM model that ameliorates the precision optimization algorithm [52], with the graphical illustration shown in Figure 5. In the local optimization part, the raw point cloud is segmented. Then one of the segments is chosen to be the reference point cloud in the ICP algorithm, and all other segments are sequentially registered by point-to-plane ICP. All the segments are merged in the end. In the pose graph construction part, the point cloud segments with high similarity are registered based on ICP as well.

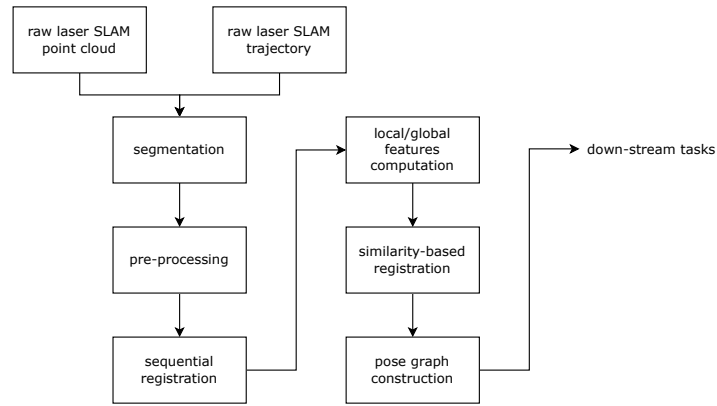

**Fig. 5.** The structure of the coarse-to-fine approach. **Left**: the local optimization part. **Right**: the pose graph construction part. This graph is rephrased from [52].

### 5.2　Landmarks in SLAM

Other than the variants according to online/offline SLAM algorithms, there are also variants of SLAM according to different environmental requirements. Typically, the environment can be divided into those with and without landmark features. Each type of mapping requires a well-designed ICP algorithm to suit with and gain a better performance.

The landmark-based SLAM utilizes landmarks detected in the input images, like trees, tall buildings, obstacles, or some artifact landmarks [53]. In the pose graph with landmarks, the nodes can represent not only the robot poses, but also the landmark locations; the edges can not only represent the odometry measurements, but also the landmark observations. An illustration of landmark mapping is shown below in Figure 6. However, there are also cases where not enough landmarks are provided [54]. If this is the case, the number of features is limited, and solutions including feature extraction and ICP algorithm are included for potential use.

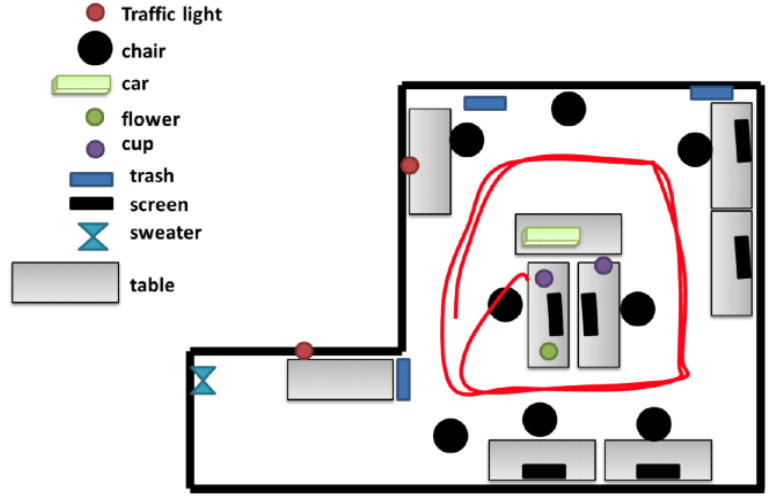

**Fig. 6.** The trajectory of the robot (red line) utilizing a landmark-based SLAM by [55]. A variety of landmarks are used in this scenario and the abundance of landmarks ameliorates the precision of the SLAM system in this research.

For landmark-based SLAM systems, they have specific requirements to ICP algorithm, because landmarks are given higher confidence value and easier to recognize. For example, Holder *et al.* (2019) proposed a real-time pose-graph-based SLAM system for automotive Radar using the ICP algorithm [53]. In this research, the ICP algorithm is used for matching consecutive scans obtained from a Radar sensor with landmarks in them with a point-to-point approach, because landmarks are also mostly constructed by points – there are few lines or planes shown in the sub-maps (Figure 7). Thus, the results still suffers from the naïve implementation of point-to-point ICP algorithm. Another research by Masahiro Tomono (2009) observed that the landmarks are typically rare in real life, so this research employs edge-points to perform as artifact landmarks in SLAM with a stereo camera [56]. The data processing part of this research utilizes an edge-point ICP variant, which projects the 3D object into a 2D one to minimize the cost function between the projected point and image edge point. This research also utilizes ICP for keyframe adjustment, which selects some keyframes to perform the ICP algorithm. Looking into integrating these two research is a valuable direction.

For non-landmark-based SLAM systems, the precision of estimation is limited due to less features, so research focuses on improving the performance to a similar degree with landmark-based SLAM systems. Cho *et al.* (2018) proposed a matching method with two sub-methods, i.e., geometric matching method and iterative closest point algorithm (Geo-ICP), as compensation for lack of landmarks [54]. In this research, the position calculated using the geometric matching method is corrected using the ICP matching method based on a point-to-line method, which means the ICP algorithm acts as a tool for decreasing error, not the usual registering process. The initial value of the ICP algorithm is calculated by the geometric matching method based on the extracted four pairs of the only shape in each map, and the correct position is estimated using the ICP matching method based on the corrected position. This attempt reminds that the ICP algorithm is essentially a statistical method, so the introduction of geometric methods may improve the performance.

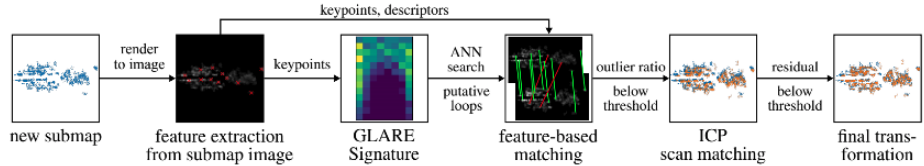

**Fig. 7.** The process of place recognition and loop closing by [53]. The feature extraction step and feature-based matching utilize landmark information, and the ICP algorithm utilizes this feature map to better perform the point-to-point approach.

## 6    Conclusion

In this paper, we reviewed the ICP algorithm and its previous taxonomy, discussed the use of ICP algorithm in SLAM tasks, and introduced a naïve taxonomy of the ICP algorithm orienting the SLAM tasks. First, the baseline ICP algorithm has a simple ideology that it sets an initial situation and iterates many times until approaching the threshold. ICP algorithm is used for 3D surface registration the most of time and has been researches for different taxonomies in previous work. Second, SLAM is developing at a rapid speed, and a certain amount of development of new SLAM systems screams for a new taxonomy of the ICP algorithm.

Our research covers the selection of ICP algorithm or Bayes filter, with the advice to use Bayes filters for state estimation problems and ICP algorithms for scan matching problems with many landmarks. Our research also points out that if we decide to use the ICP algorithm, we should determine our SLAM-oriented ICP taxonomy in two ways:
- From the criterion whether the SLAM is online or offline, ICP can be divided into time-consuming but high-precision, and timesaving but low-precision, based on different requirements of SLAM. ICP may also be applied to a subset of steps.
- From the criterion of whether the SLAM has landmark information or not, ICP can be divided into point-to-point or point-to-line/plane approach with manually creating landmarks when processing data.

# 7 Acknowledgments

I would like to acknowledge Prof. Hao Li[4], who imparted me with the most basic understandings of polygon mesh processing, 3D scanning techniques, 3D shape registration, and the ICP algorithm. I would also like to thank Prof. Cyrill Stachniss[5] for providing great open courses on YouTube and made me learn the fundamentals about SLAM.

---

[4] https://hao-li.com
[5] https://www.ipb.uni-bonn.de/people/cyrill-stachniss/